\documentclass{article}

\PassOptionsToPackage{numbers, compress}{sty/natbib}

\usepackage{sty/iclr,times}

\usepackage[utf8]{inputenc} 
\usepackage[T1]{fontenc}    
\usepackage[colorlinks=true,citecolor=.,linkcolor=.,citecolor=.,filecolor=.]{hyperref}
\usepackage{url}            
\usepackage{inconsolata}
\usepackage{booktabs}       
\usepackage{amsfonts}       
\usepackage{amssymb}
\usepackage{nicefrac}       
\usepackage{microtype}      
\usepackage{xcolor}         
\usepackage{graphicx}
\usepackage{amsmath}
\usepackage{amsfonts,amssymb}
\usepackage{multirow}
\usepackage{booktabs}
\usepackage{hyperref}
\usepackage{float}
\usepackage{enumitem}
\usepackage{algorithmicx}
\usepackage{algpseudocode}
\usepackage{tabularx}
\usepackage{subcaption}
\usepackage{marvosym}
\usepackage{textgreek}
\usepackage{wrapfig}
\usepackage{textcomp}
\usepackage{fontawesome}
\usepackage{algorithm}
\usepackage{amsmath}
\usepackage{setspace}
\usepackage{tcolorbox} 
\tcbuselibrary{skins, breakable} 

\usepackage{amsmath,amsfonts,bm}

\def\eqref#1{equation~\ref{#1}}

\def\1{\bm{1}}

\DeclareMathAlphabet{\mathsfit}{\encodingdefault}{\sfdefault}{m}{sl}
\SetMathAlphabet{\mathsfit}{bold}{\encodingdefault}{\sfdefault}{bx}{n}

\newcommand{\DS}{DS\textsubscript{T}}
\newcommand{\DU}{DU\textsubscript{T}}

\definecolor{darkblue}{rgb}{0, 0, 0.5}
\hypersetup{citecolor=darkblue}

\title{\textsc{LongAttn}: Selecting Long-context Training Data via Token-level Attention}

\author{%
  Longyun Wu~\thanks{Equal contribution.}~~$^{\heartsuit}$\quad 
  Dawei Zhu~\footnotemark[1]~~$^{\heartsuit}$  \quad
  Guangxiang Zhao~\thanks{Primary mentor}~~$^\diamondsuit$ \quad 
  Zhuocheng Yu~$^{\heartsuit}$ \\
  {\bf Jungfeng Ran~$^{\heartsuit}$ \quad Xiangyu Wong~$^{\heartsuit}$ \quad Lin Sun~\thanks{Corresponding authors.}~~$^{\diamondsuit}$  \quad Sujian Li~\footnotemark[3]~~$^{\heartsuit}$} \\
  $^\heartsuit$~Peking University \quad $^\diamondsuit$~Qiyuan Tech \\
  \texttt{wulongyun@stu.pku.edu.cn} \quad \texttt{sunlin1@360.cn} \quad \texttt{lisujian@pku.edu.cn} \\
}
\iclrfinalcopy 

\begin{document}

\maketitle

\vspace{-0.7cm}
\begin{center}
    \url{https://github.com/Lyun0912-wu/LongAttn}
\end{center}
\vspace{0.2cm}
\vspace{5pt}

\begin{abstract}
With the development of large language models (LLMs), there has been an increasing need for significant advancements in handling long contexts. To enhance long-context capabilities, constructing high-quality training data with \textbf{long-range dependencies} is crucial.
Existing methods to select long-context data often rely on sentence-level analysis,
which can be greatly optimized in both performance and efficiency.
In this paper, we propose a novel token-level framework,  \textbf{LongAttn}, which leverages the self-attention mechanism of LLMs to measure the long-range dependencies for the data. By calculating token-level dependency strength and distribution uniformity of token scores, LongAttn effectively quantifies long-range dependencies, enabling more accurate and efficient data selection. We filter \textbf{LongABC-32K} from open-source long-context datasets (ArXiv, Book, and Code). Through our comprehensive experiments, LongAttn has demonstrated its excellent \textbf{effectiveness}, \textbf{scalability}, and \textbf{efficiency}. To facilitate future research in long-context data, we released our code and the high-quality long-context training data LongABC-32K.
\end{abstract}

\section{Introduction}
Large language models (LLMs) have achieved impressive performance across a broad spectrum of traditional natural language processing tasks \citep{touvron2023llama}. To effectively address real-world applications, these models further require enhanced capabilities in handling longer contexts, particularly in key areas such as in-context learning \citep{brown2020language}, real-world question-answering based on lengthy documents \citep{wang2024leave}, long-context dialogue with historical context \citep{packer2023memgpt}, and comprehensive document summarization \citep{koh2022empirical}. 

To enhance LLMs' long-context processing capabilities, data engineering remains fundamental. Simple methods to construct long-context datasets are through naive methods like concatenating short texts or randomly sampling existing sources (e.g., CommonCrawl, GitHub). However, studies by \citet{longcontextblog} and \citet{chen2024long} emphasize that data obtained through such approaches fail to effectively improve long-context capabilities of LLMs because the data lack meaningful long-range dependencies. Inspired by this, a line of studies focus on exploring the identification and selection of high-quality long-context with consideration of relations between text segments were proposed.
ProLong \citep{chen2024long} measures long-range dependencies between segments based on the relative perplexity and relative distance. 
\citet{lv2024longwanjuan} develop a set of metrics including complexity, coherence, and cohesion  based on various  kinds of text segments (i.e., sliding windows, sentences, paragraphs) to measure the quality of long texts. However, these methods have two main drawbacks: (1) Linguistic metrics do not fully align with the underlying mechanisms of LLMs, as they often fail to capture fine-grained token-level relationships. (2) They are computationally expensive and inefficient. For example, ProLong reports that the speed of a 7B parameter model is roughly 1/16 of that of a 350M parameter model, making such methods challenging to scale for LLMs.
\begin{figure*}[t]
     \centering
      \begin{subfigure}[b]{0.58\textwidth}
         \centering
         \includegraphics[width=\textwidth]{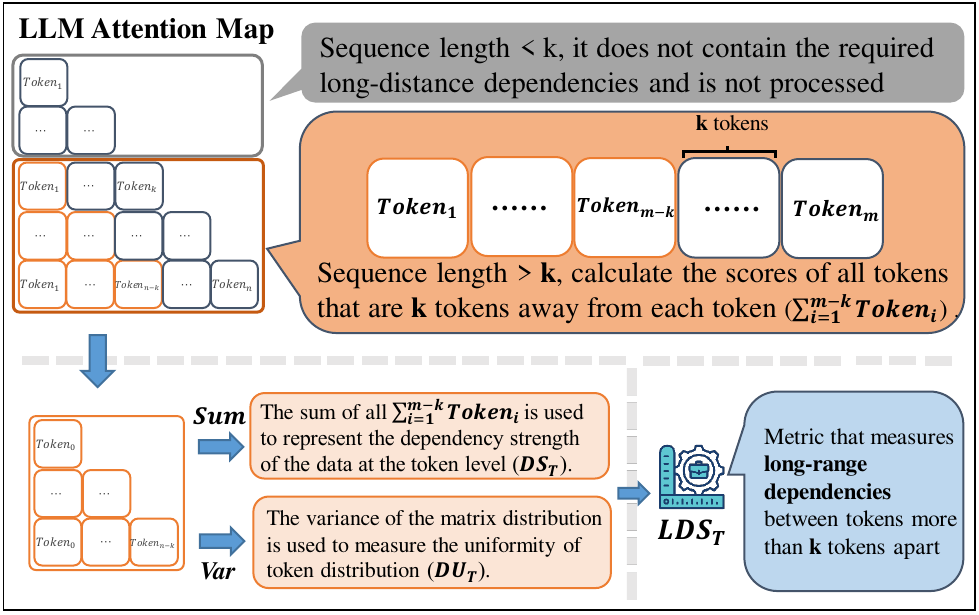}
         \caption{}
         \label{fig:method}
     \end{subfigure}
     \hfill
     \begin{subfigure}[b]{0.4\textwidth}
         \centering
         \includegraphics[width=\textwidth]{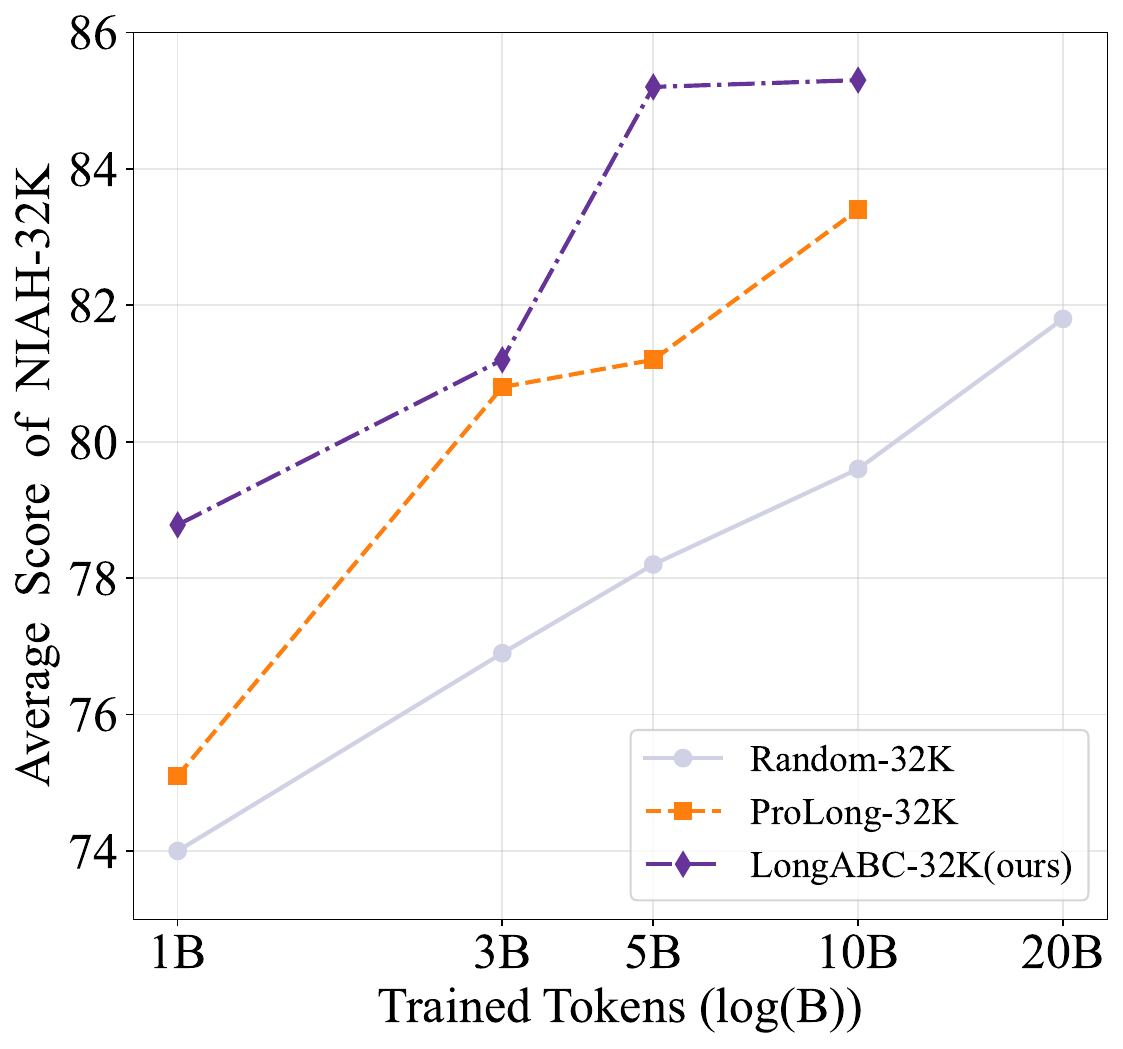}
         \caption{}
         \label{fig:niah}
     \end{subfigure}
     \caption{\textbf{(a)} How to measure long-range dependencies at the token level by using the self-attention mechanism. $DS_T$ indicates that the tokens in this data have strong long-distance dependencies, while $DU_T$ prevents negative impacts from individual tokens' high scores. \textbf{(b)} The comparison of long-context retrieval capabilities of models trained with different scales of tokens selected randomly, with sentence-level ProLong, and with LongAttn (ours).}
     \label{fig:intro_fig}
\end{figure*}

Attention mechanisms have been proven to effectively model context understanding \citep{beltagy2020longformer,zaheer2020big}. Some studies focusing on attention mechanisms and positional encoding have shown that they can significantly improve a model's long-context ability \citep{ntk,peng2023yarn}. Motivated by this, we propose to address the limitations of sentence-level selection methods leveraging the rich information provided in the attention mechanism. Specifically, we propose LongAttn, a simple yet effective framework that leverages the attention patterns of LLMs to analyse token-level dependency for long-context data selection.

LongAttn utilizes the long-range dependency indicator, $LSD_T$, to measure the strength of dependencies between tokens separated by a distance of at least $k$, which we define as \textbf{the minimum token distance}.
We break down the indicator into two scores: dependency strength($DS_T$) and distribution uniformity($DU_T$). As shown in Figure \ref{fig:method}, $DS_T$ measures the strength of dependencies between tokens separated by a distance of at least $k$, 
and $DU_T$ serves as a correction term, ensuring a consistent distribution of token scores and preventing individual tokens with excessively high attention scores from skewing the overall dependency assessment. To enhance computational efficiency and avoid the \textbf{Attention Sink} \citep{xiao2023efficient}, we use the attention score calculated by the first decoder layer of LLaMA \citep{dubey2024llama}. To better integrate 
$DS_T$ and $DU_T$, we normalize them to the same value range and then multiply the distribution uniformity ($DU_T$) by a correction factor $\alpha$. In this way, our framework effectively quantifies the degree of contextual information aggregation at the token level, providing a reliable criterion for selecting high-quality long-context data.

Through comprehensive experiments, the LongAttn framework has demonstrated significant advantages. We selected \textbf{A}rxiv, \textbf{B}ook, and \textbf{C}ode as the long-context datasets to be studied. After pre-processing, we used the LongAttn framework to make selections, and the resulting data is referred to LongABC-32K. Datasets selected using the ProLong framework \citep{chen2024long} and random selection mechanism are designated as ProLong-32K and Random-32K, respectively. As shown in Figure \ref{fig:niah}, we compare the long-context retrieval abilities of models trained on these datasets across different token scales. The experimental results demonstrate that models trained on LongABC-32K consistently perform the best, even surpassing those trained on 20B tokens from the randomly selected dataset, despite using only 5B tokens. Through further experiments, we found that, in addition to its \textbf{effectiveness} (As seen in \ref{sec:result}), LongAttn exhibits excellent \textbf{scalability} (Performs better with attention map from larger models, as seen in \ref{sec:scalability})  and \textbf{efficiency} (As seen in \ref{sec:efficiency}). Our contributions are summarized as follows:

\begin{itemize}[leftmargin=*, nolistsep]
\setlength{\itemsep}{1mm}
    \item We propose LongAttn, a framework which is the first to analyze long-range dependencies at the token level by using self-attention mechanisms.
    \item To facilitate future research in long-context data, we release LongABC-32K, a high-quality long-context dataset with strong long-range dependencies. 
    \item Through comprehensive experiments, we have demonstrated LongAttn's excellent effectiveness, scalability, and efficiency.
\end{itemize}

\section{Related Work}
\paragraph{Long-context LLMs} The ability to process extensive contextual information is a crucial aspect of language models, with context length serving as a key determinant of their processing capacity. During the pre-training phase, methods to enhance long-context capabilities primarily involve increasing the training window through Adjusted Base Frequency (ABF) and then training with selected high-quality long-context data \citep{dubey2024llama,xiong-etal-2024-effective,chen2024long,lv2024longwanjuan}. In the post-training phase, there are still efforts dedicated to post-training data \citep{gao2024train,fu2024dataengineeringscalinglanguage,si2025gateauselectinginfluentialsamples,wang2024bootstrap,Chen2024WhatAT,longalign,wu2024long}. There are also efforts dedicated to making structural adjustments, such as modifying positional encoding \citep{Chen2023ExtendingCW,zhu2023pose,peng2023yarn,ding2024longrope,an2024training,an2024does} and attention mechanisms~\citep{an2024does,jin2024llm}, aiming to more efficiently enhance the model's ability to process long contexts. Accurately assessing a model's ability to process long contexts has also become increasingly important, and a series of comprehensive and complete evaluation schemes have subsequently been proposed \citep{hsieh2024ruler,bai2023longbench, bai2025longbenchv2,kuratov2024babilong,li2024long,zhu-etal-2024-longembed,levy2024same}. From the above, it is evident that data is always crucial. Below are related works on data.

\paragraph{Pre-training data} 
Training data that exhibits long-range dependency patterns is crucial for enhancing the model's ability to handle extended contextual information. For post-training data, numerous methodologies have been explored to generate synthetic long-context data \citep{wang2024bootstrap,Chen2024WhatAT,longalign,wu2024long}. Conversely, for pre-training data, the predominant approach involves the curation and selection of relevant text from existing corpora, which is exemplified by prominent models including Qwen \citep{bai2023qwen} and LLaMA \citep{touvron2023llama}. While scaling laws suggest that a model's capabilities improve with more data \citep{kaplan2020scaling}, large volumes of data bring about high resource demands. Therefore, optimizing data utilization more effectively should become a key area of research. ProLong \citep{chen2024long} proposes a framework for calculating \textbf{long-distance dependencies} of data at the sentence level. LongWanjuan \citep{lv2024longwanjuan} also designed metrics and filtered data based at the sentence level.
However, \citet{xiong-etal-2024-effective} assert that the key factor affecting the long-context ability of LLMs is the positional encoding's capacity to aggregate information from distant tokens. Our method focuses on token-level long-distance dependencies to select high-quality long-context data.
\begin{figure*}
  \includegraphics[width=1\linewidth]{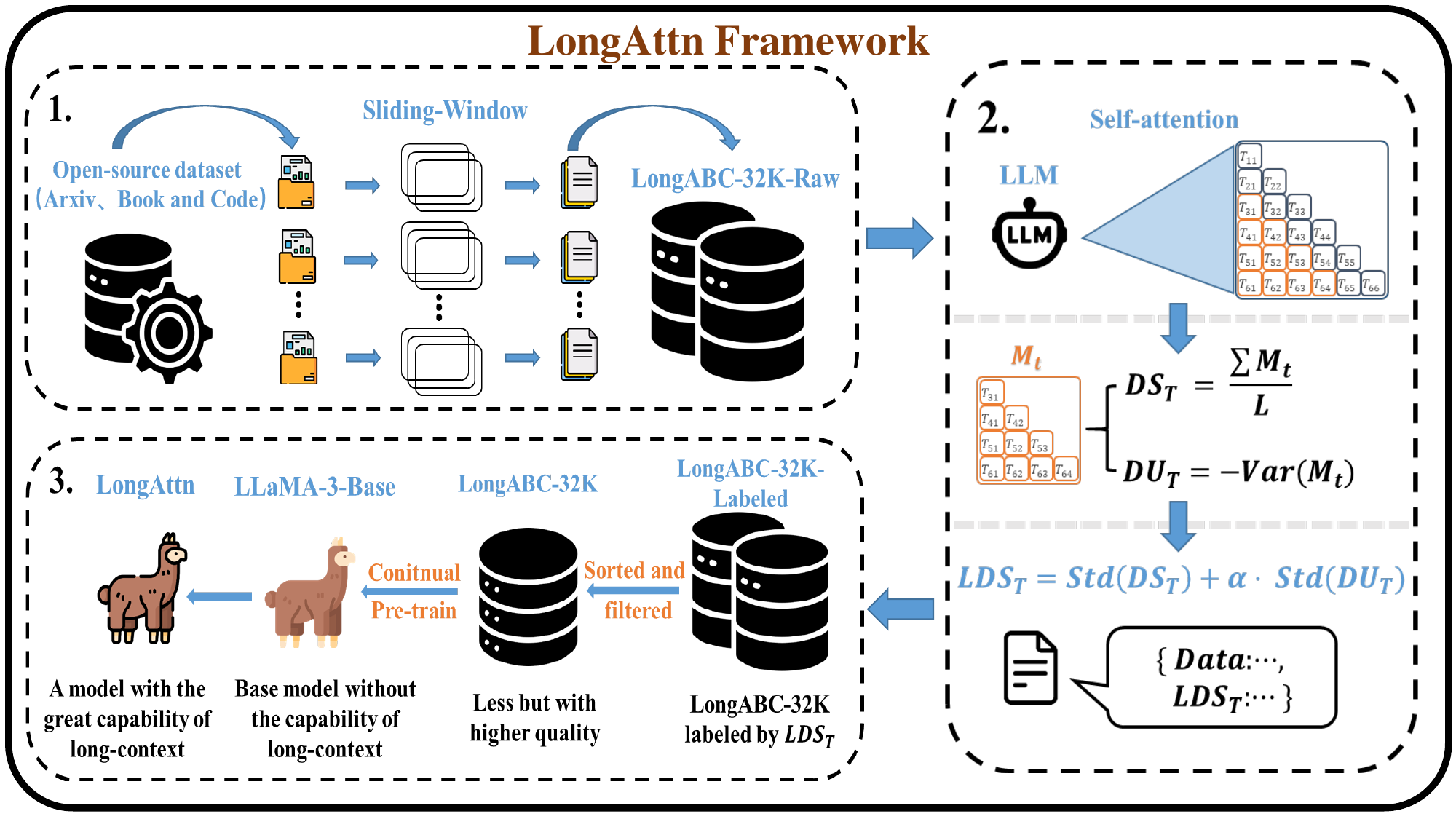}
  \caption{LongAttn Framework: After preprocessing the data, the long-distance dependency strength at the token-level is analyzed using the self-attention mechanism of an LLM. This analysis serves as the basis for filtering the data, which is then used for continual pre-training of a base model that initially lacks long-context capabilities, resulting in our LongAttn model}
  \label{fig:main}
\end{figure*}

\section{Methodology}
As shown in Figure \ref{fig:main}, our proposed method can be divided into three steps. Firstly, we gather and preprocess the data to a predetermined length. Subsequently, we employ the self-attention mechanism of a LLM to compute the long-distance dependency score for each data instance. Finally, we filter the data based on the score and utilize the refined dataset for continued pre-training of the model.

\subsection{Data Collection and Preprocessing}
\label{sec: preprocess}
To ensure the training data is suitable for long-context modeling, we carefully curate and preprocess our dataset. We choose books, code, and Arxiv papers as our primary sources of long-context data, drawing from open-source pre-training datasets such as RedPajama \citep{weber2024redpajama} and Dolma \citep{soldaini2024dolma}. These sources are known for their rich content and long sequences, which are essential for training models with extended context windows.

Given that the computational complexity of self-attention layers grows quadratically with sequence length, we set the context length to 32k tokens in this work. This length strikes a balance between capturing long-range dependencies and maintaining reasonable computational complexity. To segment/divide the data into 32k-token chunks/segments, we employ a sliding-window approach, which is more effective than naive truncation in preserving the integrity of the information. Let the total number of tokens in a text be $n$. The sliding-window strategy is as follows:
\begin{itemize}[leftmargin=*, nolistsep]
\setlength{\itemsep}{1mm}
    \item  If $32768(32k)<n\leq 65536(64k)$, take both the front and back windows. 
    \item  If $65536(64k)<n \leq 98304(96k)$, take the front, back, and middle windows. 
    \item  If $n>98304(96k) $, iteratively take the front and back windows until one of the two conditions above is met. 
\end{itemize}
The detailed algorithm is presented in Appendix \ref{sec:algorithm}. After preprocessing, we obtain the long-context pre-training dataset \textbf{LongABC-32K-Raw}, which we denote as $\mathcal{D}$. 

\subsection{Assess Long-distance Dependency via Token-level Attention}
To effectively select high-quality long-context data, we need to accurately measure the long-range dependencies within the data. In this section, we detail the process of assessing long-distance dependencies in the data using token-level attention mechanisms.
\subsubsection{Token-level Dependency Strength}
Given a data instance $s\in\mathcal{D}$, we input it into an LLM and extract the masked self-attention matrix $M$ from the first transformer decoder layer to quantify the long-range dependencies within the data. The choice of using the first layer is driven by two primary reasons: \textbf{(1)} It is computationally efficient, requiring approximately 1/32 of the inference time; \textbf{(2)} Due to the Attention Sink phenomenon \citep{xiao2023efficient}, deeper layers of the model tend to disproportionately focus on the initial tokens, irrespective of their semantic relevance to the language modeling task. Consequently, leveraging the shallow layers of the model's decoder is more optimal for capturing the contextual dependencies among tokens in the data. Define $A_{m,n}$ as the cumulative attention score assigned by $n$ to the first $m$ tokens (i.e., tokens from position 1 to $m$):
\begin{equation}
\label{eq1}
A_{m,n} = \sum_{i=1}^m{M_{i,n}}
\end{equation}
where $M_{i,n}$ represents the attention score assigned by the $n$-th token to the $i$-th token. Since the self-attention matrix $M$ has been normalized by the softmax function, it follows that $A_{n,n}=1$. For the $n$-th token in the data, where $n > k$, $A_{n-k,n}$ represents the sum of attention scores of all tokens located at least $k$ positions ahead of it. We define the contextual  dependency strength of the $n$-th token as:
{\begin{equation}
DS_T^n = \frac{A_{n-k,n}}{A_{n,n}} = A_{n-k,n}   
\end{equation}}
which quantifies the proportion of attention scores assigned to tokens at least $k$ positions prior to the $n$-th token, relative to the total attention scores. For cases where $n \leq k$, we define $DS_T^n = 0$ to account for insufficient context. Finally, the token-level contextual dependency strength of the entire data instance is defined as the average of $DS_T^n$ over all tokens:
{
\begin{align}
    DS_T &= \frac{1}{L}\sum_{i=1}^L DS_T^i \\
    &= \frac{1}{L}\sum_{i=k+1}^L DS_T^i \\
    &= \frac{1}{L}\sum{M_t}
\end{align}
}
where $L$ is the total number of tokens in the data and $M_t$ represents the lower triangular matrix in the bottom left corner of matrix $M$:
\begin{equation}
M_t = 
\begin{pmatrix}
M_{k+1,1} & 0       & \cdots & 0 \\
M_{k+2,1} & M_{k+2,2} & \cdots & 0 \\
\vdots  & \vdots  & \ddots & \vdots \\
M_{L,1} & M_{L,2} & \cdots & M_{L,L-k}
\end{pmatrix}
\end{equation}

\subsubsection{Distribution Uniformity of Token Scores} 
While $DS_T$ provides a measure of dependency strength, it is important to ensure that individual tokens with high scores do not disproportionately influence the overall dependency assessment. For example, In the previously mentioned Attention Sink phenomenon, the first token's scores very high in deeper decoder self-attention layers, which can have a significantly negative impact. Instead, the scores across the entire data segment should be consistently high. To achieve this, we introduce the distribution uniformity of token scores $DU_T$ to measure the uniformity of the score distribution:
\begin{equation}
    \label{eq4}
    DU_{T} = - Variance(M_t)
\end{equation}
This correction term helps to prevent individual tokens with excessively high attention scores from skewing the overall dependency assessment.

\subsubsection{Collaborative Ensemble}
To obtain a comprehensive measure of long-range dependencies, we combine the dependency strength $DS_T$ and the distribution uniformity $DU_T$.
Due to the differences in the magnitudes of $DU_{T}$ and $DS_{T}$, as shown as Appendix \ref{sec:Distribution}, we compute $DU_{T}$ and $DS_{T}$ for all data and then standardize them to independent normal distributions. We then use the following formula to calculate the final long-distance dependency score:
\begin{equation}
    LDS_{T} = Std(DS_{T})+\alpha \cdot Std(DU_{T})
    \label{eq: ldst}
\end{equation}
where $\alpha$ is a correction factor that balances the contributions of $DS_T$ and $DU_T$, and $Std$ represents Z-Score standardization.

\section{Experimental Setup}
\subsection{LongAttn Setup and Training Details}
In the process of filtering data using LongAttn, we utilize the first transformer decoder layer of LLaMA-3.1 to calculate long-distance dependency score. The length of each data segment $L$ is 32768 and the minimum token distance $k$ is set to $L/4$ (i.e., 8192). We set the correction factor $\alpha$ in the Eq.\ref{eq: ldst} to 0.5. 

We adopt Adjusted Base Frequency (ABF) \citep{xiong-etal-2024-effective} to continual pretrain LLaMA-3, extending the context window size to 32,768 by adjusting the RoPE theta parameter. The continued pre-training is based on the Megatron training framework \citep{shoeybi2019megatron}, utilizing 8x8 H800 GPUs. Detailed parameters can be found in the Appendix \ref{sec:Training Parameters}. 

\subsection{Continual Pre-trained Datasets}
We form the following datasets by combining short-context data with selections made through random sampling, the ProLong framework, LongAttn based on LLaMA-3.1-8B, and LongAttn based on LLaMA-3.1-70B: $\mathcal{D}_{Rx}(x\in \{1,3,5,10,20\})$, $\mathcal{D}_{Px}(x\in \{1,3,5,10\})$, $\mathcal{D}_{Ax,8B}(x\in \{1,3,5,10\})$, and $\mathcal{D}_{Ax,70B}(x\in \{1,3,5,10\})$, with $x$ representing the data size in Billions.

To ensure the diversity of the filtered data, we apply the filtering process within each category of datasets separately. For detailed data composition, please refer to the Appendix \ref{sec:Dataset}.

\subsection{Baselines}
\paragraph{Data-Scale Comparison} To demonstrate the effectiveness of LongAttn, We conduct a data-scale comparison of the long-context retrieval capabilities of models continued pre-trained on $\mathcal{D}_{Rx}(x\in \{1,3,5,10,20\})$, $\mathcal{D}_{Px}(x\in \{1,3,5,10\})$, $\mathcal{D}_{Ax,8B}(x\in \{1,3,5,10\})$, and $\mathcal{D}_{Ax,70B}(x\in \{1,3,5,10\})$.

\paragraph{Fixed-Scale Method Comparison} To demonstrate the superiority of LongAttn, we conduct fixed-data method comparison of the models trained on $\mathcal{D}_{Rx}(x\in \{5,10,20\})$,  $\mathcal{D}_{P5}$, $\mathcal{D}_{A5,8B}$, and $\mathcal{D}_{A5,70B}$. Additionally, we compare them with similarly sized models that have excellent long-context capabilities. Details of the baselines can be found in Appendix \ref{sec:baselines}.

\subsection{Evaluation Tasks} 
We assess the capability of the base model, continually pre-trained within the current window length, based on the following long-context and short-context criteria: \textbf{(1)} The best reflection of the base model's long-context capabilities is its long-context retrieval ability, followed by its performance on other long-context tasks. 
\textbf{(2)} No degradation in short-context performance.  The evaluation tasks can be divided into the following parts:

\paragraph{Long-context Retrieval} Retrieval ability is the most crucial and best reflects the model's long-context ability before post-training. The `Needle In A Haystack' task analysis in-context retrieval ability of long-context LLMs. The original `needle in a haystack' task was relatively simple. RULER \citep{hsieh2024ruler} introduced a more detailed and complex `needle in a haystack' task, and we use RULER with a length of 32k to comprehensively evaluate long-context retrieval ability. 
\paragraph{Long-context Benchmark} In addition to retrieval ability, we also want to evaluate the model's performance on formal long-context tasks. LongBench \citep{bai2023longbench} is the first proposed bilingual long-context benchmark, which includes a total of 21 tasks categorized into 6 main types, with task lengths ranging from about 0 to 20k. RULER provides longer, variable-length evaluations across 13 complex tasks. Here, we will evaluate the tasks at the 32k length to assess changes in the model's long-context capabilities.
\paragraph{Fundamental Abilities of LLMs.} 
We use HumanEval \citep{chen2021evaluating} to assess code evaluation capability and OpenBookQA \citep{mihaylov2018can} to assess book knowledge extraction ability. Additionally, we use Hellaswag\citep{zellers2019hellaswag} and MMLU \cite{hendrycks2020measuring} to assess its broader short-context fundamental capabilities.

\section{Experimental Results}
\label{sec:result}
We validate the \textbf{effectiveness}, \textbf{scalability}, and \textbf{high efficiency} of LongAttn through a series of comprehensive experiments conducted on both varying data scales and fixed data scales. 
\subsection{Performance on Retrieval Ability}
We evaluate the retrieval capabilities of models trained with LongAttn-selected data and compare them with models trained on randomly selected data and ProLong-selected data. The results are shown in Table \ref{tab:niah-ruler}. The models trained with LongAttn-selected data consistently outperform those trained on randomly selected or ProLong-selected data across all data scales, demonstrating the effectiveness of LongAttn in improving data quality for long-context modeling.

Notably, models trained on a smaller amount of data filtered using our method even outperform those trained on a larger amount of randomly selected data in retrieval tasks. For example, the model trained on just 5B tokens filtered by LongAttn outperforms models trained on 10B or even 20B randomly selected tokens. This indicates that LongAttn can significantly enhance the efficiency of data usage for long-context pre-training.

\begin{table*}[t]
\vspace{-1em}
\centering
\footnotesize
\renewcommand{\arraystretch}{1.1}
\setlength\tabcolsep{4pt}

\begin{tabular}{lcccccccccc}
\toprule

\multirow{2.5}{*}{\textbf{Method}} & \multirow{2.5}{*}{\textbf{Tokens}} & \multicolumn{3}{c}{\textbf{Niah-Single}} & \multicolumn{3}{c}{\textbf{Niah-Multikey}} & \multirow{1.5}{*}{\textbf{Multi-}} & \multirow{1.5}{*}{\textbf{Multi-}} & \multirow{1.5}{*}{\textbf{Avg.}} \\
\cmidrule(lr){3-5} \cmidrule(lr){6-8}
&  & \textbf{Sigle-1} &  \textbf{Sigle-2} & \textbf{Sigle-3} & \textbf{MK-1} & \textbf{MK-2}& \textbf{MK-3} & \textbf{Value} & \textbf{Query} &\textbf{Score} \\ \midrule
Random & \multirow{4}{*}{1 B} & 99.8 & \underline{\textbf{100.0}} & 93.4  & \underline{\textbf{91.0}} & 11.6 & 11.4  & \underline{\textbf{91.7}}          & 93.2                 & 74.0         \\
ProLong & & 99.4 & 99.8 & 92.4 & 89.2 & 10.8  & 24.0     & 91.6          & \underline{\textbf{93.6}}                & 75.1         \\
LongAttn-8 & & \underline{\textbf{100.0}} & \underline{\textbf{100.0}} & 91.4 & 88.6 & 16.2  & 19.2   & 90.3    & 93.4        & 74.9 \\
LongAttn-70 & & \underline{\textbf{100.0}} & \underline{\textbf{100.0}} & \underline{\textbf{95.4}} & 88.0 & \underline{\textbf{29.0}}  & \underline{\textbf{35.0}}   & 90.4      & 92.4        & \underline{\textbf{78.8}}\\\midrule
Random & \multirow{4}{*}{3 B} & \underline{\textbf{100.0}}   & \underline{\textbf{100.0}}    & 86.2    & 92.8    & 62    & 8.6  & 70.0   &95.9     & 76.9 \\
ProLong & & \underline{\textbf{100.0}}   & 99.8          & 79.6    & \underline{\textbf{93.8}}    & \underline{\textbf{60.4}}   & \underline{\textbf{32.0}} & 85.9    & 95.0      & 80.8\\
LongAttn-8 & & \underline{\textbf{100.0}}   & \underline{\textbf{100.0}}    & 88.8    & 92.2    & 60.0    & 31.2 & 79.7      & 94.7           & 80.8 \\
LongAttn-70 & & \underline{\textbf{100.0}}   & \underline{\textbf{100.0}}    & \underline{\textbf{91.6}}    & 93.6    & 57.4    & 19.8   & \underline{\textbf{88.8}}          & \underline{\textbf{96.2}}        & \underline{\textbf{80.9}}\\\midrule
Random   & \multirow{4}{*}{5 B}    & \underline{\textbf{100.0}}   & 99.8           & 81.8    & \underline{\textbf{94.8}}    & 56.4    & 11.8 & 84.4     & 96.5    & 78.2 \\
ProLong    & & \underline{\textbf{100.0}}   & \underline{\textbf{100.0}}    & 78.0    & 92.8    & 64.8    & 40.4  & 77.8     & 95.9       & 81.2\\
LongAttn-8 & & \underline{\textbf{100.0}}   & 99.8           & 81.6    & 92.4    & 62.6    & 37.2 & \underline{\textbf{87.6}}    & \underline{\textbf{97.3}}      & 82.3\\
LongAttn-70 & & \underline{\textbf{100.0}}   & \underline{\textbf{100.0}}    & \underline{\textbf{83.8}}    & 92.8    & \underline{\textbf{84.8}}    & \underline{\textbf{46.8}} & 78.8       & 95.2      & \underline{\textbf{85.2}} \\\midrule
Random & \multirow{4}{*}{10 B} & \underline{\textbf{100.0}}   & \underline{\textbf{100.0}}    & 84.0    & 92.6    & 58.2    & 14.2 & 90.9  & \underline{\textbf{96.9}}        & 79.6 \\
ProLong    & & \underline{\textbf{100.0}}   & \underline{\textbf{100.0}}    & 83.4    & 92.8    & 74.4    & 32.2  & 88.7     & 95.5    & 83.4 \\
LongAttn-8 & & \underline{\textbf{100.0}}   & \underline{\textbf{100.0}}           & \underline{\textbf{87.4}}    & \underline{\textbf{93.0}}    & 72.2    & 23.0 & \underline{\textbf{93.1}}     & 96.8       & 83.2\\
LongAttn-70 & & \underline{\textbf{100.0}}   & \underline{\textbf{100.0}}    & 86.8    & 92.4    & \underline{\textbf{80.6}}    & \underline{\textbf{34.4}} & 92.0               & 96.5        & \underline{\textbf{85.3}}\\\midrule
Random  & \multirow{1}{*}{20 B} & 100.0   & 100.0    & 84.6    & 91.0    & 66.2    & 22.4 & 93.3                  & 96.5                  & 81.8\\
\bottomrule
\end{tabular}
\caption{Models trained with different methods for selecting varying scales of tokens were evaluated on complex NIAH tasks. Random, ProLong, LongAttn-8, and LongAttn-70 represent random selection, selection based on the ProLong framework, selection based on LongAttn with LLaMA-3.1-8B, and selection based on LongAttn with LLaMA-3.1-70B, respectively. And \underline{\textbf{bold}} number is used to highlight the better-performing models within each data size category.}
\label{tab:niah-ruler}
\end{table*}

\subsection{Performance on Long-context Benchmark}
\begin{figure}[h]
     \centering
     \begin{subfigure}[b]{0.24\textwidth}
         \centering
         \includegraphics[width=\textwidth]{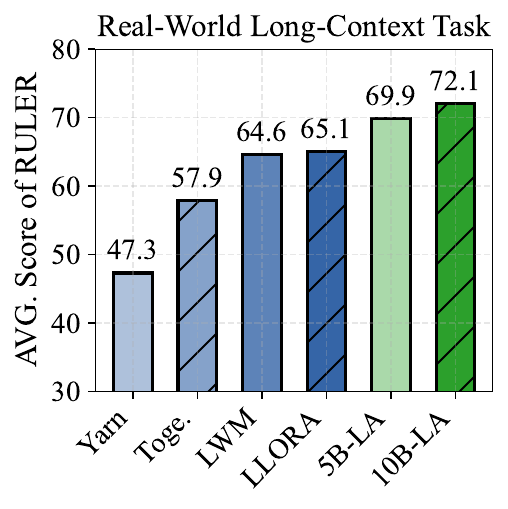}
         \caption{}
         \label{fig:others-ruler}
     \end{subfigure}
     \begin{subfigure}[b]{0.24\textwidth}
         \centering
         \includegraphics[width=\textwidth]{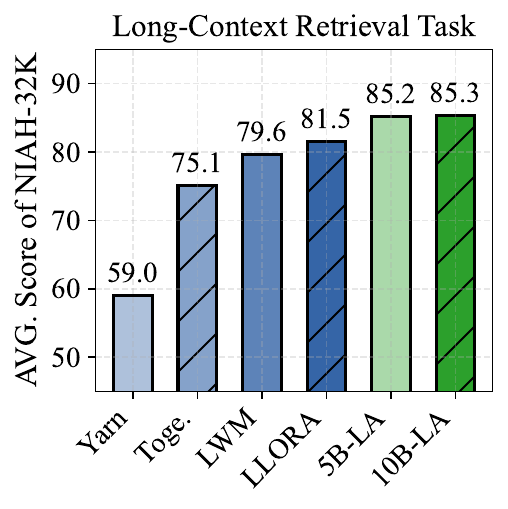}
         \caption{}
         \label{fig:others-niah}
     \end{subfigure}
     \hfill
         \centering
      \begin{subfigure}[b]{0.24\textwidth}
         \centering
         \includegraphics[width=\textwidth]{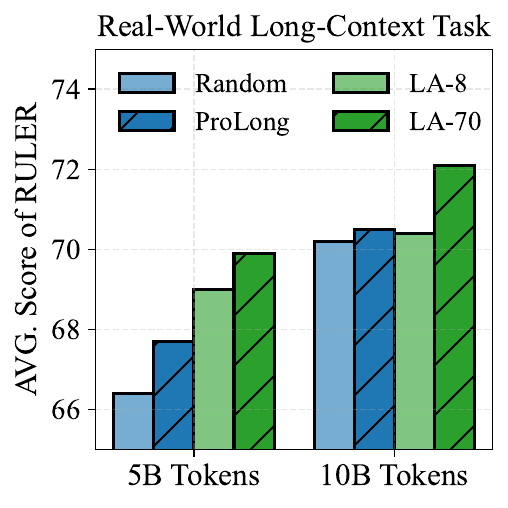}
         \caption{}
         \label{fig:ours-ruler}
     \end{subfigure}
     \hfill
     \begin{subfigure}[b]{0.24\textwidth}
         \centering
         \includegraphics[width=\textwidth]{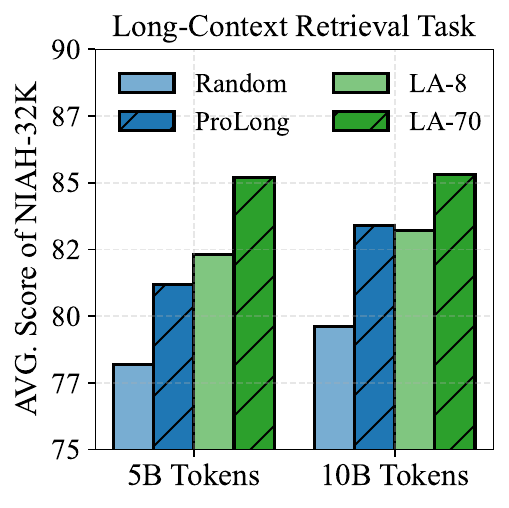}
         \caption{}
         \label{fig:ours-niah}
     \end{subfigure}
     \caption{\textbf{(a)} and \textbf{(b)} show the performance of other long-context LLMs and LongAttn-trained models on the RULER and complex NIAH tasks. \textbf{(c)} and \textbf{(d)} show the performance of models trained with different methods on the same tasks. Toge. and LLORA represent Together and LongLORA, respectively. 5B-LA and 10B-LA represent models trained on 5B and 10B tokens selected by LongAttn. LA-8 and LA-70 represent LongAttn based on 8B and 70B models, respectively.}
     \label{fig:barplot}
\end{figure}

As shown in Figure \ref{fig:others-ruler} and \ref{fig:ours-ruler}, models trained on data filtered by LongAttn outperform those trained on equivalent amounts of data selected randomly or by ProLong. LongAttn's performance is also comparable to models trained on larger data volumes. Additionally, on the RULER-32K benchmark, LongAttn outperforms all other long-context models of similar parameter sizes. The specific experimental results can be found in Appendix \ref{sec:ruler}.

As shown in Table \ref{tab: LongBench}, we compare model performance on LongBench, which consists of 21 evaluation tasks. We calculate the average score for each of the six categories to represent overall performance. The results show that LongAttn outperforms models trained on equivalent data selected randomly or by ProLong in almost all tasks and even surpasses models trained on larger amounts of randomly selected data. However, while 5B data selected by LongAttn-70 outperforms 10B randomly selected data, it does not perform as well as 5B data selected by LongAttn-8. We speculate this is because the average context length in LongBench is far below 32k, thus not effectively showcasing the advantage of 5B data selected by LongAttn-70.

\begin{table*}[h]
\vspace{-1em}
\centering
\footnotesize
\renewcommand{\arraystretch}{1.1}
\setlength\tabcolsep{4pt}

\begin{tabular}{lcccccccc}
\toprule

 \multirow{2.5}{*}{\textbf{Method}} & \multirow{2.5}{*}{\textbf{Toknes}}&\multirow{1}{*}{\textbf{Single-Doc}} & \multirow{1}{*}{\textbf{Multi-Doc}} & \multirow{1}{*}{\textbf{Summri-}} & \multirow{1}{*}{\textbf{Few-shot}} & \multirow{1}{*}{\textbf{Synthetic}} & \multirow{1}{*}{\textbf{Code Com-}} & \multirow{1}{*}{\textbf{Avg.}} \\
& & \textbf{QA} & \textbf{QA} & \textbf{zation} & \textbf{Learning} & \textbf{Tasks} & \textbf{pletion} & \textbf{Score} \\ \midrule
\multicolumn{9}{c}{\textit{Trained on 5B Tokens from Different Methods}} \\ \midrule
    Random  & 5B & 10.11              & 6.57                  & 13.72                     & 64.10                 & 1.83          & 65.05     &24.46\\
    ProLong & 5B & 11.95               & \textbf{12.59}        & 17.87                     & 63.33                 & 4.15          & 65.01     &26.93 \\
    LongAttn-8& 5B & \textbf{13.01}      & 11.20                 & 18.96            & \textbf{64.62}        & \textbf{5.12} & \textbf{65.06} &\textbf{27.46} \\
    LongAttn-70& 5B & 12.39      & 9.33                 & \textbf{19.72 }          & 64.1        & 3.42    & 65.03  &26.78 \\
    \midrule
\multicolumn{9}{c}{\textit{Trained on over 5B Tokens Selected Randomly}} \\ \midrule
    Random & 10B & 9.41         & 8.93       & 19.30         & 63.89         & 4.83     & 65.57   &26.27 \\
    Random & 20B & 11.45        & 11.72        & 20.41           & 64.13       & 9.67     & 66.51  &28.23 \\ 
\bottomrule
\end{tabular}
\caption{The performance of models continued pre-trained using data filtered by different methods on LongBench. Random, ProLong, LongAttn-8, and LongAttn-70 represent data selected randomly, data selected using the ProLong framework, data selected by the LongAttn framework with LLaMA-3.1-8B, and data selected by the LongAttn framework with LLaMA-3.1-70B, respectively. }
\label{tab: LongBench}
\end{table*}

\begin{wraptable}{r}{0.49\textwidth}
\vspace{-14pt}
\centering
\setlength\tabcolsep{3pt}
\scalebox{0.8}{
\begin{tabular}{l|l|ccccc}
\toprule
\multirow{2.5}{*}{\textbf{Model}} & \multirow{1.5}{*}{\textbf{Trained}} & \multicolumn{4}{c}{\textbf{Short-Context Task}} & \multirow{2.5}{*}{\textbf{Avg.}}  \\ 
 \cmidrule(r){3-6}
 & \textbf{Dataset} & MMLU & HS & HE & OBQA  & \\ \midrule 
 & \dag{} & \textbf{65.9} & 49.9 & 25.0 & 72.0 & 53.2 \\ 
\multirow{2.5}{*}{LLaMA} & $\mathcal{D}_{R5}$& 61.8 & 52.4 & 19.5 & 81.8 & 53.9 \\
\multirow{2.5}{*}{-3-Base} &$\mathcal{D}_{P5}$ & 61.0 & 38.3 & 23.2 & 79.4 & 50.5\\ 
 &$\mathcal{D}_{A5,8B}$ &  61.6 & 47.1 & 25.6 & \textbf{82.6} & 54.2 \\
 &$\mathcal{D}_{A5,70B}$ & 61.0 & \textbf{52.8} & \textbf{28.1} & 80.4 & \textbf{55.6} \\
\bottomrule
\end{tabular}
}
\caption{The fundamental capabilities of our continued pre-trained models and LLaMA-3-base. \dag{} indicates no training. MMLU, HS, HE, and OBQA stand for the MMLU, HellaSwag, HumanEval, and OpenBookQA tasks, respectively.}

\label{tab:fundamental task}
\end{wraptable}
\subsection{Performance on Fundamental Abilities}
The results in Table \ref{tab:fundamental task} indicate that data selected by LongAttn not only maintains the model's short-context capabilities but enhances them in specific domains. For example, the LongABC-32K-Raw dataset includes book and code data, and our model performs well on short-context tasks such as OpenBookQA \citep{mihaylov2018can} and HumanEval \citep{chen2021evaluating}.

However, there is a slight decline in performance on MMLU \cite{hendrycks2020measuring}. This is expected, as we do not include such data during continual pre-training, so the base model experienced some forgetting in these areas.

\section{Analysis}
\begin{wraptable}{r}{0.49\textwidth}
\vspace{-25pt} 
\centering
\footnotesize
\scalebox{1}{
\setlength\tabcolsep{5pt}
\begin{tabular}{cc}
\toprule
\multirow{2}{*}{\textbf{Model}} & \multirow{2}{*}{\textbf{RULER-NIAH-32K}} \\
 & \\ \midrule
$LongAttn_{\mathcal{D}_{A3}}$ & \textbf{80.83}   \\
    \quad w/ \quad$\alpha=1$ &  79.49(-1.34)            \\
    \quad w/o\quad $DU_T$    &  78.28(-2.55) \\ \midrule
$LongAttn_{\mathcal{D}_{A5}}$ & \textbf{82.30} \\
    \quad w/ \quad$\alpha=1$ & 81.05(-1.25) \\
    \quad w/o \quad$DU_T$ &  82.11(-0.19)  \\
\bottomrule
\end{tabular}
}
\caption{Ablation experiments on the constraint factor $\alpha$ and the correction term $DU_T$ were conducted on the RULER-NIAH-32K task.}
\label{tab: ablation}
\end{wraptable}
\subsection{Ablation Study}
To investigate the impact of the constraint factor $\alpha$ and the correction term $DU_T$ on regulating $LDS_T$, we conduct ablation experiments on the $\mathcal{D}_{A3}$  and $\mathcal{D}_{A5}$  datasets using retrieval tasks. The default setting of the constraint factor $\alpha$ is 0.5. 

As shown in the table \ref{tab: ablation}, we can see that the correction term $DU_T$ plays a positive role in the data selection results. In addition, the constraint on the dependency strength $DS_T$ by $DU_T$ should not be too large, which suggests that the constraint on $DS_T$ by $DU_T$ should be moderate to avoid over-correction.

\subsection{The Scalability of LongAttn}
\label{sec:scalability}
Figures \ref{fig:ours-ruler} and \ref{fig:ours-niah} show that LongAttn significantly improves performance when using stronger models. This indicates that more powerful models can better analyze the dependencies between long-context tokens. It can be envisioned that using LongAttn with larger models could yield even stronger performance.

However, in works like ProLong, computational efficiency is constrained by the  approach, making it unfeasible to use larger models. This unique advantage of LongAttn highlights its tremendous growth potential.

\begin{wraptable}{r}{0.45\textwidth}
\vspace{-18pt}
\centering
\footnotesize
\renewcommand{\arraystretch}{1.1}
\setlength\tabcolsep{4pt}
\begin{tabular}{ccc}
\toprule
\multirow{2}{*}{\textbf{Method}} & \multirow{2}{*}{\textbf{Model}} & \multirow{2}{*}{\textbf{GPU Hours}} \\
 & & \\ \midrule
    \multirow {2}{*}{ProLong} & OPT-350M &  30  \\
    & LLaMA-3.1-8B & 600\\ \midrule
    \multirow {2}{*}{LongAttn} & LLaMA-3.1-8b &  50  \\
    & LLaMA-3.1-70b & 100\\
\bottomrule
\end{tabular}
\caption{Compared the GPU hours used by different methods on LongABC-32K-Raw, using H800 GPUs. For implementation simplicity, we used the traditional attention computation method in LongAttn. If efficient methods like Flash-attn were adopted, the speed would further improve.}
\label{tab: efficiency}
\end{wraptable}

\subsection{The Efficiency of LongAttn}
\label{sec:efficiency}
Compared to sentence-level methods like ProLong, LongAttn is significantly more efficient. ProLong divides the data into sentence segments and calculates the relative perplexity and distance between each segment, which is computationally expensive, especially for LLMs. As a result, only smaller models are used in their work. In contrast, LongAttn only requires a single inference pass to obtain relative scores between all tokens, using just the first layer of the LLM's decoder. This approach is far more efficient and scalable.

Table \ref{tab: efficiency} compares the GPU hours consumed by the two methods using models of different sizes on the LongABC-32K-Raw dataset. LongAttn, even with the traditional attention computation method, is much faster than ProLong. If more efficient methods like Flash-attention were adopted, the speed of LongAttn could be further improved.

\section{Conclusion}
In this paper, we introduce LongAttn, a framework evaluates long-range dependencies at the token level. LongAttn is effective as the self-attention mechanism captures relationships between all token contexts during inference. This approach to measuring long-range dependencies aligns better with the underlying operating principles of LLMs. 
We validate the effectiveness, scalability, and high efficiency of LongAttn through a series of comprehensive experiments. Additionally, our research contributes to the previously limited study of high-quality long-context training data.
This finding suggests promising directions for future research, and we anticipate further advancements in this domain through subsequent investigations.

\section*{Limitations}

Although LongAttn has demonstrated satisfactory performance, there is still room for improvement. Specifically, we used the traditional attention map calculation method, which is inefficient. While its efficiency is satisfactory, there is still significant potential for enhancement. In future work, we hope to overcome the shortcomings, refine our method further, and advance the development of long-context capabilities in LLMs.

\bibliography{bib/custom}
\bibliographystyle{bib/iclr}

\newpage
\appendix
\section{Algorithm for Pre-process}
\label{sec:algorithm}

\begin{algorithm}[h]
\begin{tcolorbox}[
    colback=green!3, 
    colframe=black!50, 
    boxrule=0.5pt, 
    arc=3pt, 
    title=Sliding Window Sample Algorithm, 
    fonttitle=\bfseries, 
    breakable 
]
\footnotesize 
\setstretch{1.2}
\caption{} 
\label{alg:sliding-window} 
\begin{algorithmic}[1]
\Require Input data $data$ and window size $W$ (where $W > 0$).
\Ensure A set of sampled windows $S$.

\Function{SlidingWindow}{$data, W$}
    \If{$\text{len}(data) < W$}
        \State \Return $\emptyset$
    \EndIf
    \State $l \gets 0$
    \State $r \gets \text{len}(data)$
    \State $S \gets \emptyset$
    \While{$r - l > 3W$}
        \State $S \gets S \cup \{data[l:l+W]\}$ 
        \State $l \gets l + W$
        \State $S \gets S \cup \{data[r-W:r]\}$ 
        \State $r \gets r - W$
    \EndWhile
    \State $\Delta \gets r - l$
    \If{$W < \Delta \leq 2W$}
        \State $S \gets S \cup \{data[l:l+W], data[r-W:r]\}$
    \ElsIf{$2W < \Delta \leq 3W$}
        \State $m \gets l + \lfloor (\Delta - W)/2 \rfloor$
        \State $S \gets S \cup \{data[l:l+W], data[m:m+W], data[r-W:r]\}$
    \EndIf
    \State \Return $S$
\EndFunction
\end{algorithmic}
\end{tcolorbox}
\end{algorithm}
The Algorithm \ref{alg:sliding-window} demonstrates how we perform sliding window pre-processing on the data. The length of the data processed using this method will remain consistent with the window size, and compared to the truncation method, this algorithm better preserves the completeness of the original information. Some of our code is based on the \citep{360-llama-factory}.

\section{Training Details}
\subsection{Training Parameters}
\label{sec:Training Parameters}
The specific experimental parameters for continual pre-training using Megatron \citep{shoeybi2019megatron} are shown in Table \ref{tab: training details}.
\begin{table}[t]
\centering
\footnotesize
\renewcommand{\arraystretch}{1.2}
\setlength\tabcolsep{20pt}
\begin{tabular}{c|ccc}
\toprule
\multirow{2}{*}{\textbf{Params}} & \multicolumn{3}{c}{Methods} \\
\cmidrule(r){2-4}
 &\textbf{Random} &\textbf{ProLong} &\textbf{LongAttn} \\ \midrule
 learning rate(lr) & 1$\times 10^{-5}$& 1$\times 10^{-5}$& 1$\times 10^{-5}$ \\
 lr decay style &cosine&cosine&cosine\\
GPUs (H800) & $8\times 8$ & $8\times 8$  & $8\times 8$   \\
mbs & 1 & 1 & 1\\
gas & 8 & 8 & 8 \\
tp size & 8 & 8 & 8 \\
pp size & 1 & 1 & 1 \\
dropout  &0.1 &0.1 &0.1\\
seq length & 32768 & 32768 & 32768 \\
\bottomrule
\end{tabular}
\caption{Parameter settings for continual pre-training by different methods based on the Megatron framework.}
\label{tab: training details}
\end{table}
\subsection{Training Dataset}
\label{sec:Training Data}
When continuing pre-training, we use the data ratios shown as Table \ref{tab: Data Proportion}, where ArXiv, Book, and Code data refer to the data selected through different methods (random selecting, based on the ProLong \citep{chen2024long} framework, or based on the LongAttn framework).

\begin{table*}[t]
\centering
\footnotesize
\renewcommand{\arraystretch}{1.2}
\setlength\tabcolsep{20pt}
\begin{tabular}{cccc}
\toprule
\multirow{2}{*}{\textbf{Types}} & \multirow{2}{*}{\textbf{length}} & \multirow{2}{*}{\textbf{Source}} & \multirow{2}{*}{\textbf{Ratio}}  \\ 
& & & \\\midrule
Wiki & Short & Dolma \citep{soldaini2024dolma} & 3\% \\ \midrule

Github & Short & Pile \citep{gao2020pile800gbdatasetdiverse} & 3\% \\ \midrule
Web & Short & Refinedweb \citep{penedo2023refinedwebdatasetfalconllm} & 4\%  \\ \midrule
ArXiv & Long & LongABC-Arxiv & 30\% \\ \midrule
Book & Long & LongABC-Book & 30\% \\ \midrule
Code & Long & LongABC-Code & 30\% \\ 
\bottomrule
\end{tabular}
\caption{The types of data and their proportions used during the continuation of pre-training. LongABC-Arxiv, LongABC-Book, and LongABC-Code refer to the types of data selected using different methods from LongABC-32K-Raw.}
\label{tab: Data Proportion}
\end{table*}

\section{Details of Continual Pre-train Dataset}
\label{sec:Dataset}
As shown as Figure \ref{tab: Data details}, LongABC-32K-Raw is a dataset obtained by sampling long-context data and then preprocessing it as mentioned in \ref{sec: preprocess}. 

LongABC-32K-Raw serves as the data source. We filter it using different methods, including random selecting, selecting based on the ProLong framework, and selecting based on the LongAttn framework. The filtered data is then combined with quantified short-context data to form our pre-training dataset, as shown in Table \ref{tab: Data Proportion}.
\begin{table}[h]

\centering
\footnotesize
\renewcommand{\arraystretch}{1.2}
\setlength\tabcolsep{20pt}
\begin{tabular}{ccc}
\toprule
\multirow{2}{*}{\textbf{Category}} & \multirow{2}{*}{\textbf{Source}} & \multirow{2}{*}{\textbf{Scale}}  \\ 
& & \\\midrule
ArXiv & ArXiv \citep{clement2019arxiv}  & 12B Tokens\\ \midrule
\multirow{2}{*}{Book} & Dolma \cite{soldaini2024dolma},& \multirow{2}{*}{12B Tokens} \\
 & RedPajama \citep{weber2024redpajama} & \\ \midrule
Code & Dolma \citep{soldaini2024dolma} & 12B Tokens\\
\bottomrule
\end{tabular}
\caption{Data source of LongABC-32K-Raw and composition of its various parts.}

\label{tab: Data details}
\end{table}

\section{Baselines}
\label{sec:baselines}
Table \ref{tab:baselines} details the models and baselines for our data-scale and fixed-data method comparison experiments.
\begin{table*}[t]
\centering
\footnotesize
\renewcommand{\arraystretch}{1.2}
\setlength\tabcolsep{6pt}
\begin{tabular}{c|c|c|c|c}
\toprule
\multirow{1.5}{*}{\textbf{Comparison}} & \multirow{2.5}{*}{\textbf{Base Model}} & \multirow{2.5}{*}{\textbf{Trained Dataset}} & \multirow{2.5}{*}{\textbf{Selected Method}} & \multirow{2.5}{*}{\textbf{Tokens}}\\ 
\textbf{Method} & & & & \\ \midrule
& & \multirow{2}{*}{$\mathcal{D}_{Rx}$} & \multirow{2}{*}{Selected Randomly} & \multirow{2}{*}{$x\in\{1B,3B,5B$,10B,20B\}}\\
 & & & & \\ 
 \cmidrule(r){3-5}
\multirow{4}{*}{Data-Scale}  & \multirow{4}{*}{LLaMA-3} & \multirow{2}{*}{$\mathcal{D}_{Px}$} & \multirow{2}{*}{ProLong} & \multirow{2}{*}{$x\in\{1B,3B,5B,10B\}$}\\& & & & \\ 
 \cmidrule(r){3-5}
 & &\multirow{2}{*}{$\mathcal{D}_{Ax,8B}$} & \multirow{2}{*}{LongAttn-8} & \multirow{2}{*}{$x\in\{1B,3B,5B,10B\}$}\\
 & & & & \\ 
  \cmidrule(r){3-5}
 & &\multirow{2}{*}{$\mathcal{D}_{Ax,70B}$} & \multirow{2}{*}{LongAttn-70} & \multirow{2}{*}{$x\in\{1B,3B,5B,10B\}$}\\
 & & & & \\ \midrule
 
 & & \multirow{2}{*}{$\mathcal{D}_{Rx}$} & \multirow{2}{*}{Selected Randomly} & \multirow{2}{*}{$x\in\{5B,10B,20B\}$}\\
 & & & &\\ 
 \cmidrule(r){3-5}
&\multirow{4}{*}{LLaMA-3} & \multirow{2}{*}{$\mathcal{D}_{Px}$} & \multirow{2}{*}{ProLong} & \multirow{2}{*}{$x\in\{5B,10B\}$}\\
 & & & &\\ 
 \cmidrule(r){3-5}
& & \multirow{2}{*}{$\mathcal{D}_{Ax,8B}$} & \multirow{2}{*}{LongAttn-8} & \multirow{2}{*}{$x\in\{5B,10B\}$}\\
\multirow{4}{*}{Fixed-Scale} & & & &\\ 
  \cmidrule(r){3-5}
\multirow{4}{*}{Method}& & \multirow{2}{*}{$\mathcal{D}_{Ax,70B}$} & \multirow{2}{*}{LongAttn-70} & \multirow{2}{*}{$x\in\{5B,10B\}$}\\
 & & & &\\ 
\cmidrule(r){2-5}
& Yarn & \multirow{2}{*}{\dag{}} & \multirow{2}{*}{\dag{}} & \multirow{2}{*}{\dag{}} \\ 
& \citep{peng2023yarn} & & &\\ 
\cmidrule(r){2-5}
& LWM & \multirow{2}{*}{\dag{}} & \multirow{2}{*}{\dag{}} & \multirow{2}{*}{\dag{}} \\ 
&\citep{liu2024world}  & & &\\
 \cmidrule(r){2-5}
& Together & \multirow{2}{*}{\dag{}} & \multirow{2}{*}{\dag{}} & \multirow{2}{*}{\dag{}} \\ 
& \citep{Together}& &  &\\
\cmidrule(r){2-5}
& LongLORA & \multirow{2}{*}{\dag{}} & \multirow{2}{*}{\dag{}} & \multirow{2}{*}{\dag{}} \\ 
 &\citep{chen2023longlora}& &  &\\
\bottomrule
\end{tabular}
\caption{The experiments compared different models and baselines. \textbf{Selected Method} indicates the method used to filter the current training set, and \textbf{Tokens} represents the number of tokens used for training. \dag{} indicates the absence of a given option. ProLong, LongAttn-8, and LongAttn-70 represent the ProLong framework, LongAttn based on LLaMA-3.1-8B, and LongAttn based on LLaMA-3.1-70B, respectively.}
\label{tab:baselines}
\hspace{-10mm}
\end{table*}

\section{Distribution of \DS{} and \DU{}}
\label{sec:Distribution}
The distribution \DS{} and \DU{} measured by LongAttn based on LLaMA-3.1-70B is shown in Table \ref{tab: Distribution}. They are distributed across different value ranges.
\begin{table}[h]
\centering
\footnotesize
\renewcommand{\arraystretch}{1.1}
\setlength\tabcolsep{2pt}
\begin{tabular}{c|cc|cc|cc}
\toprule
\multirow{2}{*}{\textbf{Statistical}} & \multicolumn{2}{c}{\textbf{Arxiv}} & \multicolumn{2}{c}{\textbf{Book}} & \multicolumn{2}{c}{\textbf{Code}}  \\
\cmidrule(r){2-3}  \cmidrule(r){4-5}  \cmidrule(r){6-7} 
\textbf{Indicators} &\textbf{\DS{} }&\textbf{\DU{}} &\textbf{\DS{}} &\textbf{\DU{}} &\textbf{\DS{}} &\textbf{\DU{}} \\ \midrule
Min Val. & 0.25 & 2.2$\times 10^{-7}$& 0.21 &1.6$\times 10^{-7}$ & 0.18 & 9.7$\times 10^{-8}$\\ \midrule
Max Val. & 0.50 & 1.8$\times 10^{-6}$ & 0.59 &4.9$\times 10^{-6}$ &0.54  &2.4$\times 10^{-6}$ \\ \midrule
Mean & 0.43 &8.5$\times 10^{-7}$ & 0.40 &4.8$\times 10^{-7}$ &0.39 & 6.1$\times 10^{-7}$\\
\bottomrule
\end{tabular}
\caption{Statistical indicators of \DS{} and \DU{} after evaluating LongABC-32K-Raw using the LongAttn framework based on LLaMA-3.1-70B}
\label{tab: Distribution}
\end{table}
\section{Other Experimental Results}
\label{sec:ruler}
The evaluation results on RULER for models trained with data selected from LongABC-32K-Raw using different methods are shown in Table \ref{tab:RULER}. 
RULER includes 13 tasks, categorized into four major types: retrieval ability, multi-hop tracking ability, information aggregation ability, and question answering ability. The retrieval ability has been thoroughly evaluated earlier, so only the average score is presented here.
\begin{table*}[t]
\vspace{-1em}
\centering
\footnotesize
\renewcommand{\arraystretch}{1.1}
\setlength\tabcolsep{6pt}

\begin{tabular}{l|c|c|c|ccc|ccc|c}
\toprule

 \multirow{2.5}{*}{\textbf{Method}} & \multirow{2.5}{*}{\textbf{Tokens}}&\multirow{1.5}{*}{\textbf{Retrival}} & \multirow{2.5}{*}{\textbf{VT}} & \multicolumn{3}{c}{\textbf{Aggregation}} & \multicolumn{3}{c}{\textbf{QA}} &\multirow{1.5}{*}{\textbf{Avg.}} \\
 \cmidrule(r){5-7} \cmidrule(r){8-10}
& & \textbf{Avg.} & & \textbf{CWE} & \textbf{FWE} & \textbf{Avg.} & \textbf{QA1} & \textbf{QA2} & \textbf{Avg.}  & \textbf{Score} \\ \midrule
\multicolumn{11}{c}{\textit{Trained on 5B Tokens from Different Methods}} \\ \midrule
Random & \multirow{4}{*}{5B} & 78.2 & 40.6 & 31.4 & 66.7 & \textbf{49.0} & 55.2 & 43.8 & 49.5 & 66.4  \\
ProLong & & 81.2 & \textbf{51.8} & 13.0 & 65.4 & 39.2 & 57.2 & 43.4 & \textbf{50.3} & 67.7  \\
LongAttn-8 & & 82.3 & 50.3 & 19.8 & 71.0 & 45.4 & 53.4 & 44.0 & 48.7 & 69.0 \\
LongAttn-70 & & \textbf{85.2} & 43.4 & 16.8 & 68.5 & 42.7 & 55.6 & 43.0 & 49.3 & \textbf{69.9}  \\ \midrule
\multicolumn{11}{c}{\textit{Trained on 10B Tokens from Different Methods}} \\ \midrule
Random & \multirow{4}{*}{10B}  & 79.6 & 48.8 & 53.3 & 74.2 & \textbf{63.7} & 55.4 & 43.6 & 49.5 & 70.2  \\
ProLong & & 83.4 & 55.1 & 19.4 & 76.8 & 48.1 & 54.6 & 44.6 & 49.6 & 70.6  \\
LongAttn-8 & & 83.2 & 52.1 & 21.8 & 77.9 & 49.9 & 54.6 & 43.8 & 49.2 & 70.4  \\
LongAttn-70 & & \textbf{85.3} & \textbf{55.6} & 31.9 & 67.4 & 49.7 & 55.4 & 44.0 & \textbf{49.7} & \textbf{72.1}  \\ \midrule
\multicolumn{11}{c}{\textit{Trained on 20B Tokens Selected Randomly}} \\ \midrule
Random & 20B & 81.8 & 47.4 & 51.9 & 87.9 & 69.9 & 51.9 & 56.0 & 46.4 & 73.0 \\
\bottomrule
\end{tabular}
\caption{The performance of models continued pre-trained using data filtered by different methods on RULER. Random, ProLong, LongAttn-8, and LongAttn-70 represent data selected randomly, data selected using the ProLong framework, data selected by the LongAttn framework with LLaMA-3.1-8B, and data selected by the LongAttn framework with LLaMA-3.1-70B, respectively. }
\label{tab:RULER}
\end{table*}
\end{document}